\documentclass[11pt]{article}
\PassOptionsToPackage{numbers,square,comma,sort&compress}{natbib}
\makeatletter
\newif\ifacl@finalcopy    \acl@finalcopytrue
\newif\ifacl@anonymize    \acl@anonymizefalse
\newif\ifacl@linenumbers  \acl@linenumbersfalse
\newif\ifacl@pagenumbers  \acl@pagenumbersfalse
\newif\ifacl@hyperref     \acl@hyperreftrue

\usepackage{xcolor}

\ifacl@linenumbers
  \usepackage[switch,mathlines]{lineno}

  \newcount\cv@tmpc@ \newcount\cv@tmpc
  \def\fillzeros[#1]#2{\cv@tmpc@=#2\relax\ifnum\cv@tmpc@<0\cv@tmpc@=-\cv@tmpc@\fi
    \cv@tmpc=1 %
    \loop\ifnum\cv@tmpc@<10 \else \divide\cv@tmpc@ by 10 \advance\cv@tmpc by 1 \fi
      \ifnum\cv@tmpc@=10\relax\cv@tmpc@=11\relax\fi \ifnum\cv@tmpc@>10 \repeat
    \ifnum#2<0\advance\cv@tmpc1\relax-\fi
    \loop\ifnum\cv@tmpc<#1\relax0\advance\cv@tmpc1\relax\fi \ifnum\cv@tmpc<#1 \repeat
    \cv@tmpc@=#2\relax\ifnum\cv@tmpc@<0\cv@tmpc@=-\cv@tmpc@\fi \relax\the\cv@tmpc@}%
  
  \AtBeginDocument{\linenumbers}

  \usepackage{etoolbox} 

  \newcommand*\linenomathpatch[1]{%
    \expandafter\pretocmd\csname #1\endcsname {\linenomath}{}{}%
    \expandafter\pretocmd\csname #1*\endcsname {\linenomath}{}{}%
    \expandafter\apptocmd\csname end#1\endcsname {\endlinenomath}{}{}%
    \expandafter\apptocmd\csname end#1*\endcsname {\endlinenomath}{}{}%
  }
  \newcommand*\linenomathpatchAMS[1]{%
    \expandafter\pretocmd\csname #1\endcsname {\linenomathAMS}{}{}%
    \expandafter\pretocmd\csname #1*\endcsname {\linenomathAMS}{}{}%
    \expandafter\apptocmd\csname end#1\endcsname {\endlinenomath}{}{}%
    \expandafter\apptocmd\csname end#1*\endcsname {\endlinenomath}{}{}%
  }

  \expandafter\ifx\linenomath\linenomathWithnumbers
    \let\linenomathAMS\linenomathWithnumbers
    \patchcmd\linenomathAMS{\advance\postdisplaypenalty\linenopenalty}{}{}{}
  \else
    \let\linenomathAMS\linenomathNonumbers
  \fi

  \AtBeginDocument{%
    \linenomathpatch{equation}%
    \linenomathpatchAMS{gather}%
    \linenomathpatchAMS{multline}%
    \linenomathpatchAMS{align}%
    \linenomathpatchAMS{alignat}%
    \linenomathpatchAMS{flalign}%
  }
\else
  \newcommand{\@LN@col}[1]{}
  \newcommand{\@LN}[2]{}
  
\fi

\PassOptionsToPackage{a4paper,margin=2.5cm,heightrounded=true}{geometry}
\RequirePackage{geometry}

\newlength\titlebox
 \twocolumn 

\def\addcontentsline#1#2#3{}

\ifacl@pagenumbers
\else
\fi

\newcommand\outauthor{%
    \begin{tabular}[t]{c}
    \ifacl@anonymize
        \bfseries Anonymous ACL submission
    \else
        \bfseries\@author
    \fi
    \end{tabular}}

\AtBeginDocument{
\def\maketitle{\par
 \begingroup
   \def\thefootnote{\fnsymbol{footnote}}
   \twocolumn[\@maketitle]
   \@thanks
 \endgroup
 \setcounter{footnote}{0}
 \let\maketitle\relax
 \let\@maketitle\relax
 \gdef\@thanks{}\gdef\@author{}\gdef\@title{}\let\thanks\relax}
\def\@maketitle{\vbox to \titlebox{\hsize\textwidth
 \linewidth\hsize \vskip 0.125in minus 0.125in \centering
 {\Large\bfseries \@title \par} \vskip 0.2in plus 1fil minus 0.1in
 {\def\and{\unskip\enspace{\rmfamily and}\enspace}%
  \def\And{\end{tabular}\hss \egroup \hskip 1in plus 2fil
           \hbox to 0pt\bgroup\hss \begin{tabular}[t]{c}\bfseries}%
  \def\AND{\end{tabular}\hss\egroup \hfil\hfil\egroup
          \vskip 0.25in plus 1fil minus 0.125in
           \hbox to \linewidth\bgroup\large \hfil\hfil
             \hbox to 0pt\bgroup\hss \begin{tabular}[t]{c}\bfseries}
  \hbox to \linewidth\bgroup\large \hfil\hfil
    \hbox to 0pt\bgroup\hss
  \outauthor
   \hss\egroup
    \hfil\hfil\egroup}
  \vskip 0.3in plus 2fil minus 0.1in
}}
}

\renewenvironment{abstract}%
  {\begin{center}\large\textbf{\abstractname}\end{center}%
    \begin{list}{}%
      {\setlength{\rightmargin}{0.6cm}%
        \setlength{\leftmargin}{0.6cm}}%
      \item[]\ignorespaces%
      \@setsize\normalsize{12pt}\xpt\@xpt
  }%
  {\unskip\end{list}}

\RequirePackage{caption}
\DeclareCaptionFont{10pt}{\fontsize{10pt}{12pt}\selectfont}
\RequirePackage{natbib}
\renewcommand\cite{\citep}  

\def\thebibliography#1{\vskip\parskip%
\vskip\baselineskip%
\def\baselinestretch{1}%
\ifx\@currsize\normalsize\@normalsize\else\@currsize\fi%
\vskip-\parskip%
\vskip-\baselineskip%
\section*{References\@mkboth
 {References}{References}}\list
 {}{\setlength{\labelwidth}{0pt}\setlength{\leftmargin}{\parindent}
 \setlength{\itemindent}{-\parindent}}
 \def\newblock{\hskip .11em plus .33em minus -.07em}
 \sloppy\clubpenalty4000\widowpenalty4000
 \sfcode`\.=1000\relax}

\def\thesourcebibliography#1{\vskip\parskip%
\vskip\baselineskip%
\def\baselinestretch{1}%
\ifx\@currsize\normalsize\@normalsize\else\@currsize\fi%
\vskip-\parskip%
\vskip-\baselineskip%
\section*{Sources of Attested Examples\@mkboth
 {Sources of Attested Examples}{Sources of Attested Examples}}\list
 {}{\setlength{\labelwidth}{0pt}\setlength{\leftmargin}{\parindent}
 \setlength{\itemindent}{-\parindent}}
 \def\newblock{\hskip .11em plus .33em minus -.07em}
 \sloppy\clubpenalty4000\widowpenalty4000
 \sfcode`\.=1000\relax}

\def\section{\@startsection {section}{1}{\z@}{-2.0ex plus
    -0.5ex minus -.2ex}{1.5ex plus 0.3ex minus .2ex}{\large\bfseries\raggedright}}
\def\subsection{\@startsection{subsection}{2}{\z@}{-1.8ex plus
    -0.5ex minus -.2ex}{0.8ex plus .2ex}{\normalsize\bfseries\raggedright}}
\def\subsubsection{\@startsection{subsubsection}{3}{\z@}{-1.5ex plus
   -0.5ex minus -.2ex}{0.5ex plus .2ex}{\normalsize\bfseries\raggedright}}
\def\paragraph{\@startsection{paragraph}{4}{\z@}{1.5ex plus
   0.5ex minus .2ex}{-1em}{\normalsize\bfseries}}
\def\subparagraph{\@startsection{subparagraph}{5}{\parindent}{1.5ex plus
   0.5ex minus .2ex}{-1em}{\normalsize\bfseries}}

\skip\footins 9pt plus 4pt minus 2pt
\def\footnoterule{\kern-3pt \hrule width 5pc \kern 2.6pt }
\labelwidth\leftmargini\advance\labelwidth-\labelsep \labelsep 5pt

\def\@listi{\leftmargin\leftmargini}
\def\@listii{\leftmargin\leftmarginii
   \labelwidth\leftmarginii\advance\labelwidth-\labelsep
   \topsep 2pt plus 1pt minus 0.5pt
   \parsep 1pt plus 0.5pt minus 0.5pt
   \itemsep \parsep}
\def\@listiii{\leftmargin\leftmarginiii
    \labelwidth\leftmarginiii\advance\labelwidth-\labelsep
    \topsep 1pt plus 0.5pt minus 0.5pt
    \parsep \z@ \partopsep 0.5pt plus 0pt minus 0.5pt
    \itemsep \topsep}
\def\@listiv{\leftmargin\leftmarginiv
     \labelwidth\leftmarginiv\advance\labelwidth-\labelsep}
\def\@listv{\leftmargin\leftmarginv
     \labelwidth\leftmarginv\advance\labelwidth-\labelsep}
\def\@listvi{\leftmargin\leftmarginvi
     \labelwidth\leftmarginvi\advance\labelwidth-\labelsep}

\belowdisplayskip \abovedisplayskip
\def\@normalsize{\@setsize\normalsize{11pt}\xpt\@xpt}
\def\small{\@setsize\small{10pt}\ixpt\@ixpt}
\def\footnotesize{\@setsize\footnotesize{10pt}\ixpt\@ixpt}
\def\scriptsize{\@setsize\scriptsize{8pt}\viipt\@viipt}
\def\tiny{\@setsize\tiny{7pt}\vipt\@vipt}
\def\large{\@setsize\large{14pt}\xiipt\@xiipt}
\def\Large{\@setsize\Large{16pt}\xivpt\@xivpt}
\def\LARGE{\@setsize\LARGE{20pt}\xviipt\@xviipt}
\def\huge{\@setsize\huge{23pt}\xxpt\@xxpt}
\def\Huge{\@setsize\Huge{28pt}\xxvpt\@xxvpt}

\ifacl@hyperref
  \PassOptionsToPackage{breaklinks}{hyperref}
  \RequirePackage{hyperref}
  \definecolor{darkblue}{rgb}{0, 0, 0.5}
  \hypersetup{colorlinks=true, citecolor=darkblue, linkcolor=darkblue, urlcolor=darkblue}
\else
  \def\href#1#2{{#2}}
  \usepackage{url}
\fi
\makeatother

\usepackage{times}
\usepackage{latexsym}
\usepackage[T1]{fontenc}
\usepackage[utf8]{inputenc}
\usepackage{microtype}
\usepackage{booktabs}
\usepackage{amsmath,amssymb}
\usepackage{graphicx}
\usepackage{multirow}
\usepackage{xcolor}
\usepackage{indentfirst}
\makeatletter
\renewcommand\bibnumfmt[1]{[#1].}
\def\thebibliography#1{\vskip\parskip%
\vskip\baselineskip%
\def\baselinestretch{1}%
\ifx\@currsize\normalsize\@normalsize\else\@currsize\fi%
\vskip-\parskip%
\vskip-\baselineskip%
\section*{References\@mkboth
 {References}{References}}\list
 {}{\settowidth{\labelwidth}{\bibnumfmt{#1}}%
 \setlength{\leftmargin}{\labelwidth}\addtolength{\leftmargin}{\labelsep}%
 \setlength{\itemindent}{0pt}}
 \def\newblock{\hskip .11em plus .33em minus -.07em}
 \sloppy\clubpenalty4000\widowpenalty4000
 \sfcode`\.=1000\relax}

\makeatother

\newcommand{\alphaOrd}{0.325}
\newcommand{\ceilR}{0.597}
\newcommand{\ceilRmse}{3.27}

\newcommand{\sbRel}{0.811}
\newcommand{\sbSqrt}{0.90}
\newcommand{\lenientMean}{7.61}
\newcommand{\strictMean}{4.34}

\newcommand{\pctSpreadSix}{63\%}
\newcommand{\labelSd}{2.40}
\newcommand{\testN}{89}

\newcommand{\numedits}{373}
\newcommand{\numeditsTierOne}{310}
\newcommand{\meanjaccard}{0.93}
\newcommand{\minjaccard}{0.75}
\newcommand{\surfR}{0.62}
\newcommand{\maxMargin}{0.670}

\newcommand{\nMetricsChallenged}{13}
\newcommand{\minMargin}{0.011}
\newcommand{\bertMargin}{0.039}
\newcommand{\simMarginLo}{0.022}

\newcommand{\partyMarginAll}{0.035}

\newcommand{\dissocRhoAll}{0.24}
\newcommand{\dissocRhoAllLo}{-0.39}
\newcommand{\dissocRhoAllHi}{0.75}
\newcommand{\nDissocAll}{13}

\newcommand{\isoDeltaRho}{0.014}
\newcommand{\nliIdentRaw}{0.780}
\newcommand{\nliIdentPass}{1.3}
\newcommand{\nRules}{12}

\newcommand{\bestRuleMargin}{0.333}

\newcommand{\judgeR}{0.755}
\newcommand{\judgeRsd}{0.054}

\newcommand{\judgeMargin}{0.370}

\newcommand{\judgeIdent}{10.00}
\newcommand{\judgeEdit}{6.69}

\newcommand{\judgeUnrel}{1.06}

\newcommand{\asymWeaken}{4.72}
\newcommand{\asymStrengthen}{8.02}

\newcommand{\repRcalTwo}{0.726}

\newcommand{\promptMarginLo}{0.421}
\newcommand{\promptMarginHi}{0.687}
\newcommand{\promptMarginRange}{0.267}

\title{\textsc{LexFlip}: A Dissociation Diagnostic for\\Legal Meaning Preservation Metrics}

\author{Gaurab Baral \\
  University of Cincinnati \\
  \texttt{baralgb@mail.uc.edu}}

\begin{document}
\maketitle

\begin{abstract}
Does a simplified legal clause still say what the original said? The checks in
current use cannot establish that it does: requiring an identical pair to score
highest and an unrelated pair lowest moves lexical overlap and legal force
together, so any monotone function of token overlap satisfies both. Our remedy
is a \emph{dissociation}, an item holding surface form fixed while legal force
moves. We release \textsc{LexFlip}, \numedits\ minimal perturbations of Quebec
statutory French that reverse legal force while preserving \meanjaccard\ of the
tokens, with a harness scoring metrics, regressors and prompted judges alike.
The seven embedding and BERTScore metrics we test spend only
$\simMarginLo$--$\bertMargin$ of their identical-to-unrelated range on such an
edit, against $\maxMargin$ for bidirectional NLI, the one family the
identical-pair check would disqualify. On \textsc{FrJudge}, against a measured
human ceiling of $r=\ceilR$, a bare length feature outscores every semantic
metric and has the lowest margin we measure.
\end{abstract}

\begin{figure}[t]\centering
\IfFileExists{figs/teaser.pdf}{\includegraphics[width=\columnwidth]{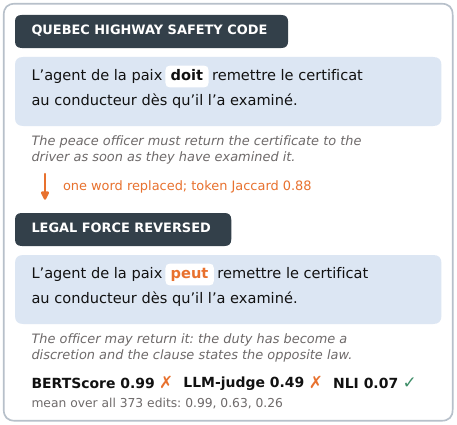}}{\fbox{\parbox{0.9\columnwidth}{\centering\vspace{2em}[figure pending]\vspace{2em}}}}
\caption{One \textsc{LexFlip} item. Replacing \emph{doit} with \emph{peut}
turns a duty into a discretion and leaves the sentence otherwise untouched, so a
metric reading legal meaning must drop and one reading overlap cannot. The
judge's ten-point rating is rescaled to $[0,1]$, as in
Table~\ref{tab:challenge}. Over the whole set the edit costs BERTScore
\simMarginLo\ of its own identical-to-unrelated range and the judge
\judgeMargin, against \maxMargin\ for bidirectional NLI.}
\label{fig:teaser}
\end{figure}

\section{Introduction}
\label{sec:intro}

Rewriting a legal text for a lay reader has an unusual failure mode. When a news
summary drops a qualifier the reader is mildly misinformed; when an insurance
clause drops a qualifier the reader may be uninsured. \emph{Moffatt v.\ Air
Canada} (2024) held an airline liable for its chatbot's misstatement of its own
policy, and legal research tools sold as hallucination-free still err on $17$ to
$33\%$ of queries \citep{magesh2025hallucinationfree}.

If simplified legal text is produced automatically, an automatic metric is what
checks it. Two checks are in current use \citep{beauchemin2023meaningbert,
beauchemin2025judgebert}: a sentence paired with itself must receive the maximum
score, and a sentence paired with an unrelated one the minimum. Both are
necessary and both are easy to state. But moving from an identical pair to an
unrelated pair changes lexical overlap and legal meaning \emph{together}, in the
same direction, at the same time, so any monotone function of token overlap
satisfies both by construction.
The checks test calibration at two endpoints; they do not test what the metric
is reading.

The remedy is a \emph{dissociation}: an item on which the two confounded
variables are pulled apart, so that legal force changes while surface form does
not. Replace \emph{doit} (must) with \emph{peut} (may) and ninety-odd percent of
the tokens are untouched, the sentence stays fluent and statutory in register,
and an obligation has become a permission (Figure~\ref{fig:teaser}). A metric
that measures legal meaning must drop sharply; a metric that measures overlap
cannot. We release
\textsc{LexFlip}, \numedits\ such perturbations of Quebec statutory French,
with a harness taking an arbitrary scoring function so that untrained metrics,
regressors and prompted judges take the same
test,\footnote{\url{https://github.com/Nyvora-Vision-Labs/LEXFLIP}} and a
four-part validation protocol around it.

\paragraph{Related work.}
BLEU does not correlate with meaning preservation once sentence splitting is
involved \citep{sulem2018bleu}, and BERTScore \citep{zhang2019bertscore} and
other embedding metrics score high across \emph{all} alterations.
\textsc{FrJudge} \citep{beauchemin2025judgebert} is to our knowledge the only
annotated resource for legal meaning preservation in any language, and trained
metrics fitted to such corpora \citep{beauchemin2023meaningbert} are our point
of departure. Closest in method, \citet{chen2023menli} show similarity metrics
are not robust to meaning-changing edits while NLI-based metrics repair much of
the gap, and \citet{mujahid2026stress} run the complementary experiment; no
prior challenge set we know of targets legal force. We follow
\citet{bean2025construct} and \citet{pacchiardi2024cleverhans} on construct
validity, \citet{deutsch2021statistical} and \citet{graham2014williams} on
statistics, \citet{xu2026disagreement} on rating distributions.

\section{What Would Validate a Legal Meaning Metric?}
\label{sec:protocol}

The four requirements below are stated for an arbitrary scorer $m(o,s)$ over an
original clause $o$ and a candidate simplification $s$; none presupposes an
architecture.

\paragraph{R1: a measured ceiling.}
A correlation with human judgment means nothing until one knows what correlation
a human achieves. Where a corpus carries $k \geq 3$ ratings per item this is
free: hold out one rater, correlate their ratings against the mean of the other
$k-1$, average over the held-out position. That answers \emph{is the metric as
good as an expert?}; the Spearman--Brown reliability of the $k$-rater mean
answers \emph{is it as good as the aggregate label allows?} A metric can sit
above the first and below the second without being superhuman, since predicting
an average is easier than being a rater.

\paragraph{R2: identical supervision.}
Comparisons here place an unsupervised cosine similarity and a fine-tuned
regressor in one table, decimal-scale the former, and report RMSE for both.
Decimal scaling fixes the range but neither the location nor the shape of the
distribution, so a similarity occupying $[0.7,1.0]$ records a large RMSE however
well it ranks, and the column measures supervision rather than judgment.
We instead fit a monotone map from raw score to human scale on the training
split alone; isotonic regression cannot improve a ranking, and rank statistics
move by at most $\isoDeltaRho$.

\paragraph{R3: a trivial-feature control.}
Simplifications are usually shorter and the dominant annotated error category is
omission, so a rubric deducting score once per identified error yields a label
partly predictable from how many words disappeared. We require a supervised
baseline over surface features alone, fitted under R2.

\paragraph{R4: dissociation.}
On naturally occurring pairs surface form and legal force vary together, the
confound the endpoint checks inherit. Let $s'$ be a perturbation of $s$ with
token Jaccard $J(s,s') \approx 1$ and legal force $L(s') \neq L(s)$, and $u$ an
unrelated sentence of the same register. We report the \emph{margin fraction}
$(\overline{m(s,s)} - \overline{m(s,s')}) /
 (\overline{m(s,s)} - \overline{m(s,u)})$, the share of the metric's own
identical-to-unrelated range spent on a legally decisive edit, which charges it
against the range it actually has rather than a nominal $[0,1]$. A metric
treating the flipped sentence as it treats an unrelated one scores $1$; a pure
overlap function scores near $0$. This is not the rate at which $(s,s)$ outranks
$(s,s')$, the weaker quantity: a metric can rank every item correctly while
moving a thousandth of its range.

\begin{table*}[tb]\centering\footnotesize
\setlength{\tabcolsep}{4pt}
\begin{tabular}{lrrrrr}
\toprule
Metric & identical & \textsc{LexFlip} & unrelated & discrim.\% & margin \\
\midrule
NLI-bidirectional (min) & 0.784 & 0.265 & 0.010 & 93.3 & 0.670 \\
NLI-bwd (no omission) & 0.784 & 0.432 & 0.042 & 77.2 & 0.473 \\
NLI-fwd (no hallucination) & 0.784 & 0.469 & 0.043 & 77.7 & 0.424 \\
LLM-judge (deepseek-chat, k=5) & 1.000 & 0.632 & 0.006 & 86.3 & 0.370 \\
surface: token Jaccard & 1.000 & 0.928 & 0.089 & 99.2 & 0.079 \\
LaBSE & 1.000 & 0.972 & 0.292 & 100.0 & 0.039 \\
BERTScore-FlauBERT & 1.000 & 0.979 & 0.393 & 100.0 & 0.035 \\
mE5-base & 1.000 & 0.975 & 0.266 & 100.0 & 0.034 \\
Para-mUSE-mpnet & 1.000 & 0.981 & 0.312 & 100.0 & 0.028 \\
Sentence-CamemBERT & 1.000 & 0.983 & 0.316 & 100.0 & 0.024 \\
BERTScore-CamemBERTv2-Recall & 1.000 & 0.983 & 0.240 & 100.0 & 0.023 \\
BERTScore-CamemBERTv2 & 1.000 & 0.982 & 0.193 & 100.0 & 0.022 \\
surface: $-|\,|s|-|o|\,|$ & 1.000 & 0.997 & 0.718 & 11.3 & 0.011 \\
\bottomrule
\end{tabular}
\caption{\textsc{LexFlip}, min--max normalised per metric. \emph{discrim.} is
the share of items ranked correctly (identical $>$ edited), \emph{margin} the
fraction of the identical-to-unrelated range spent on the legal edit.}
\label{tab:challenge}
\end{table*}

\section{\textsc{LexFlip}}
\label{sec:lexflip}

Source sentences come from two Quebec statutes held out from every corpus used
here, the Automobile Insurance Act and the Highway Safety Code, and
each takes one minimal edit that changes legal force while leaving the sentence
intact. The inventory is organised by legal effect, not by linguistic category.
Six families make up \emph{tier 1}, where the edit unambiguously changes who is
bound, what is covered or how much: modality (\emph{doit} $\leftrightarrow$
\emph{peut}), party (\emph{l'assur\'e} $\leftrightarrow$ \emph{l'assureur}),
polarity (dropping \emph{ne \dots pas}), quantum ($\times 10$), scope
(\emph{tous les} $\rightarrow$ \emph{certains}) and temporal (\emph{avant}
$\leftrightarrow$ \emph{apr\`es}); \emph{tier 2} connective and class shifts are
reported separately. The diagnostic would fail if a metric could find the edited
item by noticing that it reads badly, so perturbations introducing a grammatical
artefact are filtered by a conservative French well-formedness check that flags
none of the $400$ unmodified sources. The result is \numedits\ pairs
(\numeditsTierOne\ tier 1) with mean token Jaccard \meanjaccard, none below
\minjaccard.

For each $s$ with perturbation $s'$ we score $(s,s)$, $(s,s')$ and $(s,u)$ for
an unrelated $u$ from the same statutes, min--max normalise per metric so the
two probes anchor the range, and take the margin from the three means. Only
French-capable metrics are administered, since scoring English-only ones on
French confounds coverage with quality. The set covers three families, named in
Table~\ref{tab:challenge}: BERTScore
\citep{zhang2019bertscore} over CamemBERTv2 and FlauBERT, in F1 and recall-only
variants since recall targets omission; four sentence-embedding cosines; and
bidirectional NLI with mDeBERTa-XNLI \citep{laurer2024nli}, where failure of
$o \Rightarrow s$ signals hallucination and of $s \Rightarrow o$ omission, the
two dominant categories of the legal error taxonomy. We add token Jaccard and
the surface features of R3.

\section{Results}
\label{sec:results}

\subsection{The Diagnostic}
\label{sec:diagnostic}

Every \emph{similarity}-based metric fails, and fails almost completely
(Table~\ref{tab:challenge}). The seven embedding and BERTScore variants spend
between $\simMarginLo$ and $\bertMargin$ of their working range on an edit that
reverses the law: BERTScore over CamemBERTv2 places a sentence saying the
opposite of the original at $0.982$, where the original sits at $1.000$ and an
unrelated statute at $0.193$. Token Jaccard, at $0.079$, is in the same regime,
which is the point: these are overlap functions, exactly as
Section~\ref{sec:protocol} says they must be. Their \emph{discrimination} rates
are uninformative for the same reason, the seven ranking the identical pair
first $100\%$ of the time by a margin that carries no decision.
Bidirectional NLI does not fail: it spends $\maxMargin$ of its range on the
edit, ranks correctly on $93.3\%$ of items, and places a flipped sentence nearer
the unrelated pair than the identical one.

\begin{table}[tb]\centering\footnotesize
\setlength{\tabcolsep}{4pt}
\begin{tabular}{lrr}
\toprule
Perturbation & $n$ & mean margin \\
\midrule
drop \emph{ne \dots pas} & 21 & 0.333 \\
\emph{avant} $\leftrightarrow$ \emph{apr\`es} & 9 & 0.272 \\
amount $\times$ 10 & 21 & 0.256 \\
\emph{doit} $\rightarrow$ \emph{peut} & 82 & 0.241 \\
\emph{peut} $\rightarrow$ \emph{doit} & 111 & 0.167 \\
\emph{tous les} $\rightarrow$ \emph{certains} & 42 & 0.153 \\
\emph{et} $\leftrightarrow$ \emph{ou} & 53 & 0.069 \\
\emph{assureur} $\leftrightarrow$ \emph{assur\'e} & 14 & 0.035 \\
\midrule
\textbf{All \nRules\ rules} & 373 & \textbf{0.155} \\
\bottomrule
\end{tabular}

\caption{Margin over the \nMetricsChallenged\ scorers of
Table~\ref{tab:challenge}; selected rules, same-direction pairs pooled. All
\nRules\ rules are in the supplementary material.}
\label{tab:rules}
\end{table}

\paragraph{The existing checks would have excluded it.}
Note where NLI's identical score sits: $0.784$, and not as a normalisation
artefact. In raw terms mDeBERTa assigns a mean entailment of $\nliIdentRaw$ to a
sentence paired with \emph{itself}, clearing the $99\%$ bar the identical-pair
convention requires on only $\nliIdentPass\%$ of items, against $100\%$ for
every BERTScore and embedding metric. The metric that best tracks legal meaning
is the one convention would \emph{disqualify}. Saturation on identical input is
easy to build, every overlap function having it by construction: report it, but
do not filter on it. Nor is any rule reliably detected
(Table~\ref{tab:rules}), the best averaging $\bestRuleMargin$ and the party swap
$\partyMarginAll$.

\begin{table*}[tb]\centering\footnotesize
\setlength{\tabcolsep}{4pt}
\begin{tabular}{lrrrrr}
\toprule
Metric & $r$ & $\rho$ & RMSE$_{\text{raw}}$ & RMSE$_{\text{cal}}$ & over\% \\
\midrule
\multicolumn{6}{l}{\emph{Unsupervised, French-capable}}\\
BERTScore-CamemBERTv2 & 0.291$_{\pm0.09}$ & 0.326 & 2.62 & 2.36 & 49.3 \\
BERTScore-FlauBERT & 0.482$_{\pm0.08}$ & 0.459 & 2.16 & 2.12 & 49.4 \\
LaBSE & 0.409$_{\pm0.06}$ & 0.402 & 2.33 & 2.21 & 47.2 \\
mE5-base & 0.358$_{\pm0.08}$ & 0.356 & 2.52 & 2.26 & 48.0 \\
NLI-bidirectional (min) & 0.395$_{\pm0.07}$ & 0.405 & 3.74 & 2.22 & 46.6 \\
\midrule
\multicolumn{6}{l}{\emph{Trivial surface features}}\\
$-|\,|s|-|o|\,|$ (length difference) & 0.641$_{\pm0.04}$ & 0.624 & 3.25 & 1.85 & 47.0 \\
Token Jaccard & 0.303$_{\pm0.06}$ & 0.331 & 3.34 & 2.34 & 48.0 \\
Surface-Ridge (10 features, supervised) & 0.621$_{\pm0.05}$ & 0.588 & 1.89 & 1.89 & 45.8 \\
\midrule
\multicolumn{6}{l}{\emph{Prompted LLM judge}}\\
LLM-judge (deepseek-chat, k=5) & 0.755$_{\pm0.05}$ & 0.753 & 2.93 & 1.57 & 43.9 \\
\midrule
\textbf{Human ceiling} (one annotator) & \textbf{0.597} & 0.620 & -- & 3.27 & -- \\
\bottomrule
\end{tabular}

\caption{Calibrated comparison on the \textsc{FrJudge} test split ($\testN$
items), mean over 10 seeds; all ten unsupervised metrics are in the
supplementary material. RMSE$_{\text{raw}}$ is the decimal-scaling protocol in
current use, RMSE$_{\text{cal}}$ the isotonic map of R2.}
\label{tab:baselines}
\end{table*}

\subsection{The \textsc{FrJudge} Corpus}
\label{sec:casestudy}

The diagnostic answers R4. We run the other three on \textsc{FrJudge}
\citep{beauchemin2025judgebert}, 297 insurance clauses simplified by
\texttt{gpt-4-turbo-2024-04-09} and rated by five law students on legal meaning
($1$--$10$), then ask whether the two kinds of evidence agree. The public
release drops annotator identity, making R1 impossible downstream, so we rebuild
the corpus from the raw Prodigy export, recovering the five annotators
(pseudonymised A--E) and reproducing the published pairwise agreement of
$25.96\%$.

\paragraph{Disagreement and the ceiling.}
Krippendorff's $\alpha$ for legal meaning is $0.104$ under the \emph{nominal}
coefficient, the one conventionally reported here, and $\alphaOrd$ under the
\emph{ordinal} one appropriate to an ordered scale
\citep{krippendorff2004content}. What remains is structural: the annotators
separate into two stable populations, B, C and D averaging $\lenientMean$
against $\strictMean$ for A and E, and \pctSpreadSix\ of pairs span at least $6$
points. Averaging leaves a target with standard deviation $\labelSd$, the
condition under which a regressor predicting the middle scores well. Correlating each annotator against the mean of the other four gives a
ceiling of $\ceilR$ (RMSE $\ceilRmse$); exceeding it is not superhuman, the
bound for predicting that mean being its own reliability, $\sbRel$ by
Spearman--Brown, so a metric can approach $\sqrt{\sbRel}=\sbSqrt$.

\paragraph{Identical supervision, and a trivial control.}
Calibration collapses most of the RMSE gap uncalibrated comparisons record
(Table~\ref{tab:baselines}): the metrics were not as far from human judgment as
such tables suggest, merely on a different scale. And a ridge over word counts
and token overlap reaches $r=\surfR$, at the human ceiling and above every
semantic metric we test, the best reaching $r=0.482$. Only two scorers beat it,
and neither is semantic: the prompted judge below, and the bare length
difference at $r=0.641$. This says less about the corpus than about what the
\emph{annotation} rewards, score being reduced once per identified error and the
most frequent error type, omission, correlating strongly with becoming shorter.

\begin{figure}[tb]\centering
\IfFileExists{figs/dissociation.pdf}{\includegraphics[width=0.66\columnwidth]{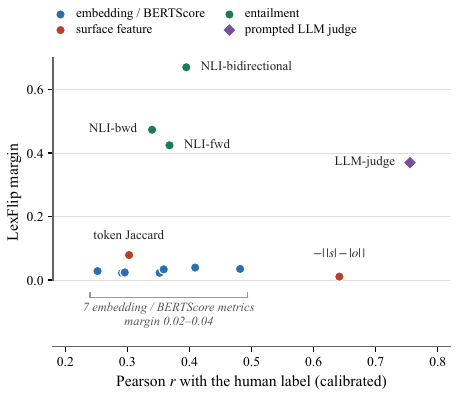}}{\fbox{\parbox{0.9\columnwidth}{\centering\vspace{2em}[figure pending]\vspace{2em}}}}
\caption{The two orderings come apart. Horizontally, calibrated correlation with
the human label, the quantity a conventional comparison selects on; vertically,
the share of working range spent on an edit that reverses the law.}
\label{fig:dissociation}
\end{figure}

\paragraph{A prompted judge.}
A prompted model given the annotators' rubric \citep{enguehard2025lemaj} is the
natural competitor to a fine-tuned 112M-parameter encoder. We score each pair
with \texttt{deepseek-chat} under the \textsc{FrJudge} scale, a ten-point
judgment deducting one point per identified legal error, in its own request so
the judge never sees the probe condition, $K=5$ times. Under R2 it
reaches $r=\judgeR_{\pm\judgeRsd}$, above the length feature and our ceiling (a
redraw gave $\repRcalTwo$). On \textsc{LexFlip} it scores the identical pair
at $\judgeIdent$ and the unrelated pair at $\judgeUnrel$, flawless on the two
checks jointly satisfiable by overlap, then awards a clause whose legal force
has been reversed $\judgeEdit$ out of ten, a margin of $\judgeMargin$. The
rubric, not the model, limits this: with the model fixed the margin moves from
$\promptMarginLo$ under the rubric-faithful prompt to $\promptMarginHi$ under a
bare instruction stripping the four-type taxonomy, against $0.002$ for a redraw.
Score is reduced \emph{once per identified error}, so a clause reversal is one
incoherence and the penalty is capped near the top of the scale. One behaviour
survives every configuration: \emph{peut} for \emph{doit} scores $\asymWeaken$
while the reverse scores $\asymStrengthen$, though both change the law by the
same construction, so the judge tracks legal \emph{severity} rather than legal
\emph{equivalence}.

\section{Discussion and Conclusion}
\label{sec:discussion}

Over the \nDissocAll\ scorers carrying both a calibrated $r$ and a margin the
rank correlation between them is $\rho=\dissocRhoAll$ ($95\%$ interval
$[\dissocRhoAllLo, \dissocRhoAllHi]$), an absence of resolution rather than a
demonstrated null; the extremes are the durable part. The length feature
outscores every semantic metric and has the lowest margin of anything we test
($\minMargin$), while bidirectional NLI sits mid-table on correlation and alone
reacts to the edit. The empty upper-right quadrant of
Figure~\ref{fig:dissociation} is the finding.

A metric claiming to measure legal meaning should therefore report three checks,
not two: identical pairs (saturation), unrelated pairs (bottoming out), and
\textsc{LexFlip} (does it move when, and only when, the law moves?). The first
two are jointly satisfiable by lexical overlap; the third is not, so a low
margin is a fixable deficiency rather than a limit of automatic evaluation. Only
legal sensitivity resists calibration: rank on it, and build the next metric on
a repaired entailment objective.

\section*{Limitations}
\label{sec:limits}

Our \textsc{LexFlip} labels are definitional rather than annotated: we assert
that replacing \emph{doit} with \emph{peut} changes legal meaning rather than
having lawyers rate each perturbation. That is defensible for tier-1 edits, reported
separately from tier 2, but a small expert study would be stronger. The edits
are single and template-generated, so a system could be tuned to them without
acquiring general legal sensitivity, which is why the diagnostic is a necessary
condition and not a benchmark to optimise.

The same caveat bears on our best result. Our perturbations are largely
negation, modality, quantifier and antonym contrasts, close to what NLI
training data is built around, so bidirectional NLI's margin of $\maxMargin$ may
reflect familiarity with their \emph{form} rather than sensitivity to their
legal consequence. \citet{yang2026compared} give the general version: a
counterfactual edit isolates the variable of interest only against a
meaning-preserving paraphrase control, which we do not run and the next version
of the diagnostic should carry.

Finally, the diagnostic and the corpus comparison run on different text,
insurance forms against two statutes, and answer different questions; the judge
results rest on one model under a rubric reconstructed from the published
description, a load-bearing assumption given that wording alone moves the margin
by $\promptMarginRange$; and the ceiling is computed on all 297 items while model
correlations run on $\testN$-item splits, so ``model $>$ ceiling'' carries the
sampling uncertainty of both.

\section*{Ethical Considerations}
\label{sec:ethics}

\textsc{FrJudge}'s five annotators are pseudonymised as A--E throughout, our
released code pseudonymises annotator identifiers on load, and we ship the
pseudonymisation script for anyone redistributing a reconstruction.
\citet{beauchemin2025judgebert} caution that their metric informs a practitioner
about legal meaning preservation rather than supplying a complete juridical
analysis; our results sharpen that into a rule. A metric that cannot distinguish
\emph{doit} from \emph{peut} must not gate whether a simplified clause is safe
to publish, and the appropriate deployment is triage, the ranking of clauses for
human review, rather than certification. Nothing here constitutes legal advice.
The perturbed sentences are deliberate misstatements of Quebec law, labelled as
such in the distribution, and are not statements of the law.

\newpage

\end{document}